\documentclass[article]{amsart}
\usepackage{amssymb,amsmath,amsthm, comment,hyperref,bm,graphicx}
\usepackage{ bbold }
\usepackage{lipsum}

\numberwithin{equation}{section}

\theoremstyle{plain}

\def\mod#1{{\ifmmode\text{\rm\ (mod~$#1$)}
\else\discretionary{}{}{\hbox{ }}\rm(mod~$#1$)\fi}}

\theoremstyle{definition}

\theoremstyle{remark}

\usepackage[symbol]{footmisc}
\renewcommand{\thefootnote}{\fnsymbol{footnote}}

\begin{document}
\title[\resizebox{4.5in}{!}{Robust Bayesian Tensor Factorization with Zero-Inflated Poisson Model and Consensus Aggregation}]{Robust Bayesian Tensor Factorization with Zero-Inflated Poisson Model and Consensus Aggregation}


\subjclass[2010]{Primary: 11E76,11P05,12D15,14N10}
\keywords{}

\maketitle{}

\def\thefootnote{*}\footnotetext{These authors contributed equally to this work}

 \hspace{.5in}\author{Daniel Chafamo} \textsuperscript{1,2}$^{*}$,
  \author{Vignesh Shanmugam} \textsuperscript{1,3,4}$^{*}$, \author{Neriman Tokcan}\textsuperscript{1,5}$^{*}$

\bigskip

  {\small \textsuperscript{1} Broad Institute of MIT and Harvard, Cambridge, MA, USA \par
  
  \textsuperscript{2}Klarman Cell Observatory, Broad Institute of MIT and Harvard, Cambridge, MA, USA \par 
  
  \textsuperscript{3}Department of Pathology, Brigham and Women’s Hospital, Boston, MA\par

\textsuperscript{4}Harvard Medical School, Boston, MA, USA \par 
\textsuperscript{5}Department of Mathematics, University of Massachusetts Boston, Boston, MA \par }

\bigskip

\begin{abstract}

Tensor factorizations (TF) are powerful tools for the efficient representation and analysis of multidimensional data. However, classic TF methods based on maximum likelihood estimation underperform when applied to zero-inflated count data, such as single-cell RNA sequencing (scRNA-seq) data. Additionally, the stochasticity inherent in TFs results in factors that vary across repeated runs, making interpretation and reproducibility of the results challenging. In this paper, we introduce Zero Inflated Poisson Tensor Factorization (ZIPTF), a novel approach for the factorization of high-dimensional count data with excess zeros. To address the challenge of stochasticity, we introduce Consensus Zero Inflated Poisson Tensor Factorization (C-ZIPTF), which combines ZIPTF with a consensus-based meta-analysis. We evaluate our proposed ZIPTF and C-ZIPTF on synthetic zero-inflated count data and synthetic and real scRNA-seq data. ZIPTF consistently outperforms baseline matrix and tensor factorization methods in terms of reconstruction accuracy for zero-inflated data. When the probability of excess zeros is high, ZIPTF achieves up to $2.4\times$ better accuracy. Additionally, C-ZIPTF significantly improves the consistency and accuracy of the factorization. When tested on both synthetic and real scRNA-seq data, ZIPTF and C-ZIPTF consistently recover known and biologically meaningful gene expression programs. All our data and code are available at: \url{https://github.com/klarman-cell-observatory/scBTF} and \url{https://github.com/klarman-cell-observatory/scbtf_experiments}.

.
\end{abstract}

\section{Introduction}
Tensors are multi-way arrays that extend matrices to higher dimensions and provide a natural way to represent multidimensional data. Traditional matrix methods \emph{matricize} tensors, limiting their ability to exploit the intrinsic multi-way structure of the data \cite{Kolda}. Tensor factorization extends matrix factorization to higher dimensions while preserving the said intrinsic structure and enabling the discovery of complex interactions within the data. Several variants of tensor factorization methods exist, among which Candecomp/Parafac~(CP) and Tucker are the most widely used \cite{tensor-survey}. Tensor factorizations have found applications in fields such as computer vision, neuroscience, genomics, recommender systems, and social network analysis~\cite{Kolda, denoising,semi-nonnegative, recommender1,  NTF-image1, tensor-survey, Monti}.

Classic tensor factorization methods using maximum likelihood estimation
(MLE) can be unreliable when applied to sparse count data \cite{tensor-sparse}. Bayesian Poisson Tensor Factorization (BPTF)---a higher-order extension of Poisson matrix factorization---is used to overcome the limitations of the MLE approach when dealing with high-dimensional count data. BPTF provides advantages such as the ability to incorporate prior knowledge, perform model selection, and quantify uncertainty in parameter estimates \cite{probabilistic, bayesian, BNTF-3}. Highly-dispersed count data with excess number of zeros is common in various fields such as healthcare (e.g., hospital readmissions), genomics (e.g., gene expression levels), social sciences (e.g., user behaviors), and insurance claims \cite{Chigona,Simchowitz}. The Zero-Inflated Poisson (ZIP) distribution is a better model for such data compared to the Poisson distribution \cite{Lambert, Chigona, Ghosh}, and has been successfully used in recommendation systems and other applications \cite{Simchowitz}.

In addition to modeling the distribution of data and noise appropriately, another issue to be addressed is the inherent randomness  of tensor factorization algorithms.~This leads to varying results for multiple runs and negatively impacts the interpretability and reproducibility \cite{Kolda, Bro}. In this paper, we propose a novel approach for stable tensor factorization which is robust for high-dimensional sparse count data with excess zeros~(Section \ref{section-BPT}). We claim three \textbf{main contributions}:
\begin{itemize}
\item We propose a novel factorization approach for high-dimensional sparse count data with excess zeros, namely {\it Zero Inflated Poisson Tensor Factorization (ZIPTF)}, which utilizes the Bayesian ZIP model (Section \ref{section-BPT}).
\item To address the discussed randomness issue, we develop a meta-analysis method that generalizes consensus matrix factorization \cite{consensus-nmf} and incorporates novel  techniques to improve the stability and interpretability of the factorization results (Section \ref{Section-consensus}, Figure \ref{fig:consensus}). We specifically focus on its integration with ZIPTF, namely {\it Consensus-ZIPTF (C-ZIPTF)}. Nonetheless, our method is generalizable to other factorization approaches.
\item We provide an extensive evaluation on three different datasets:~(1)~synthetic zero-inflated count tensors with increasing probability $\Phi$ of excess zeros (Section \ref{section:synthetic_tensor});~(2)~synthetic multi-sample single-cell RNA sequencing (scRNA-seq) data (Section \ref{section:simulated});~(3)~real scRNA-seq dataset of immune cells stimulated with interferon beta (Section \ref{section:single-cell}).
We compare ZIPTF and C-ZIPTF against baseline matrix and tensor factorization methods. Our results indicate that ZIPTF outperforms the baselines in terms of reconstruction accuracy for zero-inflated data. Specifically, for $\Phi=0.8$, ZIPTF achieves an average explained variance of 0.92, compared to a maximum of 0.38 achieved by the baseline models. Additionally, C-ZIPTF significantly improves the consistency and accuracy of the factorization results. Finally, both ZIPTF and C-ZIPTF successfully capture biologically meaningful gene expression programs (GEPs) and result in factors with higher Pearson correlations to known GEPs.
\end{itemize}

\section{Tensor preliminaries}
This section presents the foundational concepts and notations for tensors, with most of the notation borrowed from \cite{Kolda}.  
We denote the $(i_1, i_2, \ldots, i_N)$-th entry of an $N$- way  tensor $\mathcal{X} \in \mathbb{R}^{I_1 \times I_2 \times \ldots \times I_N}$ as $\mathcal{X}_{i_1i_2\ldots i_N}$. The Frobenius norm of a tensor is similar to the matrix Frobenius norm:
\begin{equation}
     \|\mathcal {X}\|_{F}= \sqrt{\sum_ {i_1=1}^{I_1}\sum_{i_2=1}^{I_2}\ldots \sum_{i_N=1}^{I_N}\mathcal{X}_{i_1i_2\ldots i_N}^2}.
\end{equation} 
An $N$-way tensor $\mathcal{Y}$ is called a rank-1 tensor if it can be written as outer product of $N$ vectors, i.e, 
$\mathcal{Y}=u^{(1)}\otimes u^{(2)} \otimes \ldots \otimes u^{(N)}$ with $\mathcal{Y}{i_1i_2\ldots i_N}= u{i_1}^{(1)}u_{i_2}^{(2)}\ldots u_{i_N}^{(N)}$.
A rank $R\geq 1$ approximation to the tensor $\mathcal{X} \in \mathbb{R}^{I_1 \times I_2 \ldots \times I_N}$ can given as:
\begin{equation}\label{decomp-cp}
    \mathcal{X} = \mathcal{\tilde{X}} + \mathcal{E}~\text{where}~ \mathcal{\tilde{X}}= \sum_{r=1}^{R} a^{(1)}_{r}\otimes a^{(2)}_{r} \otimes \dots \otimes a^{(N)}_{r},
\end{equation}
$A^{(i)}=[a^{(i)}_{1}  \ldots a^{(i)}_{R} ] \in \mathbb{R}^{I_i \times R}, 1\leq i \leq N$ is the \textit{factor matrix} along the $i-$th mode, and $\mathcal{E} \in \mathbb{R}^{I_1 \times I_2   \ldots \times I_N}$.  The factorization given in Eqn. \eqref{decomp-cp} is often referred to as the CP (Candecomp / Parafac) decomposition which is a special case of Tucker decomposition (see \cite{Kolda} for details). The approximation can be concisely expressed as $\tilde{\mathcal{X}}=[[A^{(1)},~A^{(2)},\ldots,~A^{(N)}]] .$
 In this paper, we impose a non-negativity constraint on factors to improve their interpretability. The primary method for solving  Eqn. \eqref{decomp-cp} involves using the maximum likelihood estimation (MLE) approach, which entails minimizing the following error:
\begin{equation}\label{tensor-ml}
\min_{A^{(1)}, A^{(2)}, \ldots, A^{(N)}} || \mathcal{X} - \tilde{\mathcal{X}}||_{F}.
\end{equation}
Iterative algorithms such as multiplicative updates, alternating least, and gradient descent are commonly utilized for Eqn. \eqref{tensor-ml} \cite{Acar, Kolda, Bro}. The MLE approach often assumes Gaussian noise \cite{Kolda, Bro}.

\section{Bayesian tensor factorization and consensus aggregation}\label{section-BPT}

\subsection{Bayesian Poisson tensor factorization}\label{BPF}
\newcommand{\dummyfig}[1]{
  \centering
  \fbox{
    \begin{minipage}[c][0.3\textheight][c]{0.5\textwidth}
      \centering{#1}
    \end{minipage}
  }
}

Traditional tensor factorization methods using MLE are unstable when applied to zero-inflated count data \cite{tensor-sparse}. Bayesian Poisson Tensor Factorization (BPTF) extends the Poisson Matrix Factorization method to higher dimensions and utilizes Bayesian inference to obtain a point estimate and offers benefits such as uncertainty quantification, realistic noise assumptions, and principled inclusion of prior information \cite{probabilistic, BNTF-3, bayesian, BTF-cemgil2}. This section presents a general framework for BPTF with Variational Inference~(VI)~for high-dimensional count data.
Let $\mathcal{X} \in \mathbb{R}^{I_1 \times I_2 \ldots \times I_N}$ be the observed count data drawn from the Poisson distribution and with the CP decomposition as given in Eqn.~\eqref{decomp-cp}.~Let $I=i_1i_2\ldots i_N \in \overline{I}=\{ i_1i_2 \ldots i_N~:~1 \leq i_j \leq  I_j,~1\leq j\leq N\}$, then
\begin{equation}\label{bayesian-cp1}
 X_I \approx Poisson (\lambda_{I})~~~\text{where}~ \mathcal{X}_{I} \approx \tilde{\mathcal{X}}_{I}= \sum_{r=1}^{R} a^{(1)}_{i_1r}  a^{(2)}_{i_2r}  \ldots  a^{(N)}_{i_{N}r}  \approx \lambda_I.
 \end{equation}
  BPTF uses Gamma priors to regularize the estimation of the latent factors \cite{Cemgil, beta, David}. The Gamma distribution, which is characterized by a shape parameter $\alpha >0 $ and a rate parameter $\alpha\beta > 0$, is employed as  a sparsity-inducing prior \cite{bayesian, probabilistic, Cemgil}. Then for each $a^{(k)}_{jr}$ in Eqn. \eqref{bayesian-cp1}, we have:
    \begin{equation}\label{gamma}
        a^{(k)}_{jr} \approx Gamma(\alpha,  \alpha\beta^{(k)}),~1\leq k \leq N,
    \end{equation}
with the expectation $E[a^{(k)}_{jr}]=\frac{1} {\beta^{(k)}}$ and $Var[a^{(k)}_{jr}]=\frac{1}{\alpha{\beta^{(k)}}^2}$. The posterior distribution given by $P(A^{(1)}, A^{(2)}, \ldots, A^{(N)} | \mathcal{X}, \mathcal{H})$ is intractable due to the inability to compute the evidence, given a model hyperparameter set $\mathcal{H} = \{\alpha, B^{(1)}, B^{(2)},\ldots, B^{(N)}\}$ \cite{David}.
 BPTF uses VI and assumes a variational family of distributions $\mathcal{Q}_V=\mathcal{Q}(A^{(1)},A^{(2)},\ldots, A^{(N)}; V^{(1)},\ldots, V^{(N)})$ which is indexed by a set of variational parameters $V^{(k)}, 1\leq k \leq N$ \cite{David, Bishop}. We employ a fully-factorized mean-field approximation assuming that $\mathcal{Q_V}(A^{(1)},A^{(2)},\ldots, A^{(N)})= \prod_{k=1}^{N} \mathcal{Q_V}(A^{(k)}; V^{(k)}),$ where 
$\mathcal{Q}(a^{(k)}_{jr}; V^{(k)}_{jr}) = Gamma(a^{(k)}_{jr}; \gamma^{(k)}_{jr}, \delta^{(k)}_{jr}), 1\leq k \leq  N.$ The variational family $Q$ used here is similar to the one employed in Bayesian Poisson Matrix Factorization \cite{Cemgil, poisson-nmf-recommender, Bayesian-inference-matrix}.
BPTF fits variational parameters by minimizing the Kullback-Leibler (KL) divergence between the true posterior distribution and $\mathcal{Q_V}$, which is equal to maximizing the evidence lower bound (ELBO) \cite{David, Bishop, probabilistic}:
\begin{equation}
    ELBO(V)= E_{Q_V}[log(P(\mathcal{X}, A^{(1)},A^{(2)},\ldots, A^{(N)} | \mathcal{H})]+ H(Q_V).
\end{equation}
where $H(Q_V)$ is the entropy for $Q_V.$ 
Coordinate ascent algorithms are commonly used to maximize the ELBO by iteratively optimizing each variational parameter while fixing the others until convergence, monitored by the relative change in the ELBO \cite{Bishop, David}. From Eqn. \eqref{bayesian-cp1}, we have the total $n= \sum_{I \in \overline{I}} \mathcal{X}_{I} \approx~ Poisson(\Lambda)$ where $\Lambda=\sum_{I \in \overline{I}} \lambda_I.$ We can use the Poisson-Multinomial connection to express $\mathcal{X}$ given $n$ as $Multinomial(n, \pi)$ where $(\pi)_{I}= \frac{\lambda_I}{\Lambda}$, and update variational parameters using this auxiliary distribution~\cite{David, bayesian, Cemgil, Kurt}:
\begin{eqnarray}\label{variational-updates1}
\gamma^{(k)}_{jr} &=&  \alpha + \sum_{\substack{i_1 i_2 \ldots i_N \in \overline{I} \\ i_k=j}} \mathcal{X}_{i_1 i_2\ldots i_n} \frac{\mathbb{G}_{Q_V} \big[\prod_{s=1}^{N} a^{(s)}_{i_sr}\big]}{\sum_{t=1}^{R} \mathbb{G}_{Q_V}\big[\prod_{s=1}^{N} a^{(s)}_{i_st}\big]}, \label{geometric}\\
\label{variational-updates2}
\delta^{(k)}_{jr} & = &   \alpha \beta^{(k)}  + \sum_{i_1 i_2 \ldots i_N \in \overline{I}} E_{Q_V} \big[ \prod_{ 1\leq s \neq k \leq N}a^{(s)}_{i_sr}~\big], \label{arithmetic}
\end{eqnarray}
where $E_{Q_V}[.]$ and $\mathbb{G}_{Q_V}=exp(E_{Q_V}[log(.)])$ denote arithmetic and geometric expectations.
Since $Q_V$ is fully factorized, the expectations in Equations \eqref{variational-updates1} and \eqref{variational-updates2} can be expressed as a product of individual expectations \cite{David}. Specifically, for $a^{(s)}_{i_sr},$
\begin{equation}
E_{Q_V}[a^{(s)}_{i_sr}]=\frac{\gamma^{(s)}_{i_sr} }{\delta^{(s)}_{i_sr}}~~\text{and}~~\mathbb{G}_{Q_V}[a^{(s)}_{i_sr}]=\frac{exp(\Psi(\gamma^{(s)}_{i_sr} ))}{\delta^{(s)}_{i_sr}},
\end{equation}
where $\Psi$ is the digamma function~(logarithmic derivative of the gamma function). An empirical Bayes approach can be used to update the hyperparameters $\beta^{(k)}, 1\leq k \leq N$, in conjunction with the variational parameters \cite{Cemgil, David}:
\begin{equation}\label{betaupdate}
    \beta^{(k)}= \big( \sum_{j=1}^{I_j} \sum_{r=1}^{R} \mathbb{E}_{Q_V}[a^{(k)}_{jr}]\big )^{-1}.
\end{equation}
The variational inference algorithm for BPTF is fully specified by the set of update equations Equations \eqref{variational-updates1}, \eqref{variational-updates2}, and \eqref{betaupdate}. 
\subsection{Zero-inflated Poisson tensor factorization (ZIPTF)}\label{section:ZIPTF}
Poisson models may not always be sufficient to model count data with excess zeros, and zero-inflated models can often provide a better fit \cite{Lambert,Simchowitz}. 
 The Zero-Inflated Poisson (ZIP) model assumes that the counts in the tensor $\mathcal{X}$ can be modeled as a mixture of a point mass at zero and a Poisson distribution with parameter $\lambda$. Let $\mathcal{X}$ be a count data in $\mathbb{R}^{I_1 \times I_2 \ldots \times I_N}.$ 
We define the index set $\overline{I}$ as the collection of all possible indices, i.e., $\overline{I}=\{ i_1i_2 \ldots i_N~:~1 \leq i_j \leq  I_j,~1\leq j\leq N\}.$ We say $\mathcal{X}$ has Zero-inflated Poisson (ZIP) distribution if for every $I \in \overline{I}:$ 
\begin{equation}\label{zip-model}
P(\mathcal{X}_{I}=x_{I})= p_I {\mathbb{1}_{x_{I}=0}}  + (1-p_I) \frac{e^{-\lambda} \lambda^{x_{I}} }{x_{I}!},
\end{equation}
where the outcome variable $x_I$ has non-negative integer values, $\lambda_I$ is the expected Poisson count, and $p_I$ is the \textit{probability of extra zeros} \cite{Lambert}. As an abbreviation, we write it as $\mathcal{X}_I \sim ZIP (\lambda_I, p_I).$ The ZIP  can be considered as the \emph{product} of a Poisson random variable $\mathcal{Y}_{I} \sim Poisson(\lambda_I)$ and an independent Bernoulli variable $\Phi_I \sim Bernoulli(p_I)$ \cite{Chigona}. The Bernoulli variable $\Phi_I$ takes the value of $1$ when $\mathcal{X}_{I}$ is equal to $0$, due to the Bernoulli component, and takes the value of $0$ otherwise.

We consider the low rank $R \geq 1$ decomposition  of the zero-inflated count tensor $\mathcal{X} :$
\begin{equation}\label{decomp-cp}
    \mathcal{X} \approx  \sum_{r=1}^{R} a^{(1)}_{r}\otimes a^{(2)}_{r} \otimes \dots \otimes a^{(N)}_{r}.
\end{equation}
Hence, for $I=i_1 i_2\ldots i_N$, the reconstruction $\sum_{r=1}^{R} a^{(1)}_{i_1r}  a^{(2)}_{i_2r}  \ldots  a^{(N)}_{i{N}r}$ can be interpreted as the mean of the distribution from which the observed count $\mathcal{X}_{I}$ is assumed to be sampled. Then we have:
\begin{equation}
    \mathcal{X}_{I} \sim ZIP (\lambda_I= \sum_{r=1}^{R} a^{(1)}_{i_1r}  a^{(2)}_{i_2r}  \ldots  a^{(N)}_{i{N}r}, p_{I}).
    \label{cp-zip}
\end{equation}

\subsection{Variational Inference for ZIPTF}
For given position $I=i_1 i_2 \ldots i_N,$ we consider the rank $R$ decomposition in Eqn. \eqref{cp-zip}. In Bayesian Poisson factorizations, the Gamma distribution is utilized as a prior to induce sparsity, and it is assumed that each latent factor matrix $A^{(k)} = [a^{(k)}_{1} \ldots a^{(k)}_{R}] \in \mathbb{R}_{+}^{I_k \times R},~1 \leq k \leq N$, follows a Gamma distribution \cite{Cemgil, bayesian}. Therefore, for each $a^{(k)}_{jr}$ in Eqn. \eqref{cp-zip}, we have:
\begin{equation}\label{gamma}
a^{(k)}_{jr} \sim Gamma(\alpha^{(k)}, \beta^{(k)}), \quad 1\leq k \leq N,
\end{equation}
where $\alpha^{(k)} >0 $ and $\beta^{(k)}> 0 $ represent the shape and rate parameters of the distribution,
with the expectation $E[a^{(k)}_{jr}]=\frac{\alpha^{(k)} }{\beta^{(k)}}$ and $Var[a^{(k)}_{jr}]=\frac{\alpha^{(k)}}{{\beta^{(k)}}^2}$.
Additionally, for ZIP models a  latent variable $\xi$ is introduced to capture the hidden state of the probability of extra zeros which specify $\Phi \sim Bernoulli (p_I)$ \cite{Simchowitz,David}. Let $S(.)$ denote the \textit{logistic sigmoid} function, given by $S(x)=\frac{1}{1 + e^{-x}}$, then:
\begin{equation}
\xi= S(\zeta)~~\text{where}~~\zeta \sim Normal(\mu, \sigma).
\end{equation}

Let $Z= \{A^{(1)}, A^{(2)}, \ldots, A^{(N)}, \Phi\},$ consider the posterior distribution $P(Z | \mathcal{X}, \mathcal{H}),$ given a model hyperparameter set $\mathcal{H} = \{\alpha^{(1)}, \beta^{(1)}, \alpha^{(2)}, \ldots, \beta^{(2)}, \ldots, \alpha^{(N)}, \beta^{(N)}, \mu, \sigma \}.$ 

Variational inference approximates the true posterior distribution using a family of probability distributions $\mathcal{Q}$ over hidden variables \cite{David}. This family of distributions is characterized by free parameters, and the key assumption is that each latent variable is independently distributed given these parameters. We assume a variational family of distributions $\mathcal{Q}$ indexed by a set of variational parameters $V=\{\gamma^{(1)}, \delta^{(1)},\gamma^{(2)}, \delta^{(2)}, \dots, \gamma^{(N)}, \delta^{(N)}, \overline{\mu}, \overline{\sigma}\}$ where $(\gamma^{(k)}, \delta^{(k)})$ are variational shape and rate parameters of the Gamma distribution for the latent factor along the $k-$th mode, and $(\overline{\mu}, \overline{\sigma})$ are the variational parameters for $\zeta$. We use a fully factorized mean-field approximation \cite{David} and the variational distribution factors as the following: 
\begin{equation}
\mathcal{Q} (A^{(1)},A^{(2)},\ldots, A^{(N)}, \Phi)=  \mathcal{Q}( \Phi; \overline{\mu}, \overline{\sigma}) \prod_{k=1}^{N} \mathcal{Q}(A^{(k)};\gamma^{(k)}, \delta^{(k)}).
\end{equation}

where $a^{(k)}_{jr}  \sim Gamma( \gamma^{(k)}_{jr},\delta^{(k)}_{jr})$ and $\Phi_{I} \sim Bernoulli(S(\zeta))~\text{for}~\zeta \sim Normal(\overline{\mu}, \overline{\sigma})$. The goal is to choose a member $q^{*}$ of the variational family variational distributions which minimizes the Kullback-Leibler (KL) divergence of the exact  posterior from $\mathcal{Q}$: 

\begin{equation}
q^{*}(Z) =  \arg\min_{q(Z) \in \mathcal{Q}} D_{KL} \left (q(Z)\middle\| P(Z| \mathcal{X}, \mathcal{H})\right)
\label{KL}.
\end{equation}

Upon examining the KL divergence, we encounter a significant challenge: it involves the true posterior distribution $P(Z| \mathcal{X}, \mathcal{H})$, which is not known. Nevertheless, we can rewrite the KL divergence as follows:
\begin{eqnarray}
D_{KL} \left(q(Z)\middle\| P(Z| \mathcal{X}, \mathcal{H} ) \right) & = & \int q(Z) \log \left(\frac{q(Z)  } {P(Z | \mathcal{X}, \mathcal{H} )} \right)dZ~~~~~~~~`\\
 &=& \int  q(Z)  \log \left(\frac{q(Z) P(\mathcal{X}, \mathcal{H} )} {P(Z,\mathcal{X}, \mathcal{H})} \right)dZ\\
 ~~&=&\log\left(P(\mathcal{X}, \mathcal{H} )\right) \int q (Z) dZ- \int q (Z) \log\left(\frac{P(Z,\mathcal{X}, \mathcal{H} )}{ q(Z)}\right)dZ~~~~~~~~\\
 &=& \log\left(P(\mathcal{X}, \mathcal{H} )\right) - \int q (Z) \log\left(\frac{P(Z,\mathcal{X}, \mathcal{H} )}{ q(Z)}\right)dZ. ~~~\label{ELBO}
\end{eqnarray}

The second term in Eqn. \eqref{ELBO} is called Evidence Lower Bound (ELBO). We know that the KL divergence is non-negative, therefore, $\log\left(P(\mathcal{X}, \mathcal{H} )\right) \geq~\text{ELBO}(q(Z))=\int q (Z) \log\left(\frac{P(Z,\mathcal{X}, \mathcal{H} }{ q (Z)}\right )dZ .$

\begin{eqnarray}
\text{ELBO}(q(Z)) & = &  \int  q(Z)  \log\left(P(Z,\mathcal{X}, \mathcal{H} )\right)dZ - \int q(Z) \log\left( q(Z)\right)dZ  \\ 
& =& E_{q(Z)} [\log\left(P(\mathcal{X},Z, \mathcal{H}\right)]-E_{q(Z)}[\log q(Z)] .
\end{eqnarray}
The evidence lower bound serves as a transformative tool that converts intractable inference problems into optimization problems that can be tackled using gradient-based methods \cite{David}. 

Coordinate ascent algorithms are frequently employed in maximizing the evidence lower bound (ELBO)\cite{David,Simchowitz}. However, these algorithms require tedious gradient calculations and may not scale well for very large datasets \cite{SVI, black-box}. Closed-form coordinate-ascent updates are applicable to conditionally conjugate exponential family models, but they necessitate analytic computation of various expectations for each new model\cite{SVI, black-box}.

Stochastic Variational Inference (SVI) \cite{SVI} offers a more efficient algorithm by incorporating stochastic optimization \cite{Robbins}. This technique involves utilizing noisy estimates of the gradient of the objective function. To maximize the evidence lower bound (ELBO), we employ a stochastic optimization algorithm known as the \textit{Black Box Inference Algorithm} \cite{black-box}. This algorithm operates by stochastically optimizing the variational objective using Monte Carlo samples from the variational distribution to compute the noisy gradient (see Section 2, \cite{black-box} for details). By doing so, it effectively alleviates the burden of analytic computations and provides a more efficient approach to ELBO maximization.

\begin{centering}%
\begin{figure*}[t]
    \includegraphics[width=\linewidth]{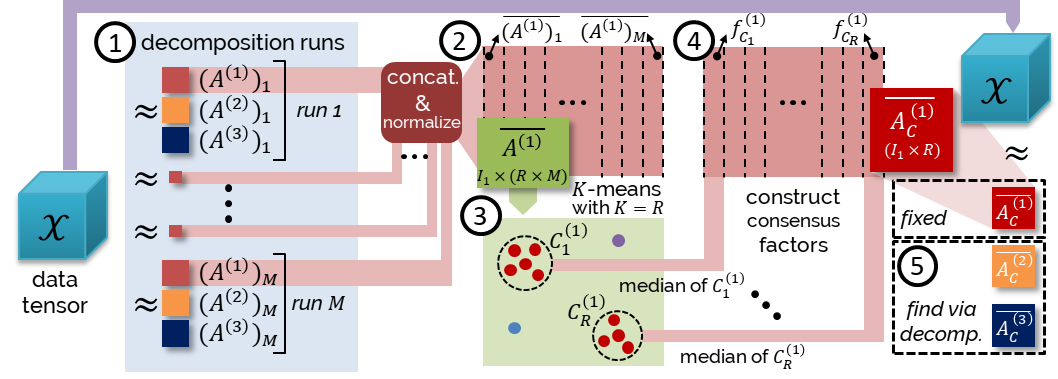}%
  \caption{Overview of the consensus meta-analysis approach discussed in Section \ref{Section-consensus} for the 3-way tensor $\mathcal{X}$.}%
\label{fig:consensus}%
\end{figure*}%
\end{centering}
\subsection{Generic consensus-based tensor factorization}\label{Section-consensus}%
Selecting the number of components in tensor factorization is challenging \cite{Kolda, Acar}. The dependence on initial guesses for latent factors can lead to substantially different factor sets across repeated runs, making it difficult to interpret the results \cite{Kolda, Acar, Bro}. We typically select the minimum value of $R$ in Eqn. \eqref{decomp-cp} that maximizes the explained variance of the approximation, defined as follows:
\begin{equation}\label{explained-variance}
\text{explained variance}=1- \frac{||\mathcal{X}- \tilde{\mathcal{X}}||_{F}}{||\mathcal{X}||_{F}}.
\end{equation}
Our goal is not solely to improve the explained variance, but also to ensure the interpretability and stability of the factors. We generalize the consensus meta-analysis approach, which has been previously used for matrix factorization \cite{consensus-nmf}, and include novel techniques to enhance the stability. The overview of the proposed pipeline is depicted in Figure \ref{fig:consensus}. In the remainder of this section, we will refer to the steps \textcircled{1}-\textcircled{5} given in the figure.

Running a generic rank $R$ factorization given~in~ Eqn.~\eqref{decomp-cp} for~ $\mathcal{X} \in \mathbb{R}^{I_1 \times I_2 \times  \ldots \times I_N}$ with $M$ different random seeds yields the sets of non-negative factor matrices $\{(A^{(1)})_{m}, (A^{(2)})_{m},\ldots, (A^{(N)})_{m} \}, ~1\leq m \leq M$~(Step \textcircled{1}). For a chosen modality $k~(1\leq k\leq N$), we can aggregate and normalize the factor matrices from independent runs (Step \textcircled{2}): 
\begin{equation}\label{aggreageted}
    \overline{A^{(k)}}=\Big[ \frac{(A^{(k)})_{1}}{||(A^{(k)})_{1}||_{F}}~\frac{(A^{(k)})_{2}}{||(A^{(k)})_{2}||_{F}}~\ldots  \frac{(A^{(k)})_{M}}{||(A^{(k)})_{M}||_{F}} \Big] \in I_k \times (R \times M).
\end{equation}
The cophenetic correlation coefficient, commonly used to select ranks for matrix factorizations \cite{Brunet}, assumes one-to-one mapping between features and factors based on maximum loadings. However, this assumption may not be valid when a feature contributes significantly to multiple factors.

Our method for selecting the rank and evaluating factorization stability involves clustering column factors of aggregated matrices and fixing the initial guess to ensure reliability.
Initially, we perform K-means clustering \cite{K-means} on the columns of the aggregated factor matrix $\overline{A^{(k)}}$ with $K=R$~(Step \textcircled{3}). The resulting cluster sets are given as $C^{(k)} _{i}= \{ \text{columns of}~\overline{A^{(k)}}~\text{assigned to cluster}~ i\}, 1\leq i \leq R.$ The Local Outlier Factor algorithm \cite{LOF} is used to remove outliers by considering the local density deviation of a data point compared to its neighbors.
We evaluate the \emph{goodness} of the clustering with the silhouette coefficient \cite{silhouette}, computed as (b-a)/max(a,b), where a is the average intra-cluster distance, and b is the average inter-cluster distance. The silhouette coefficient ranges from -1 to 1, with higher values indicating more coherence.
After clustering, we obtain the consensus factors $f^{(k)}_{C_i}$, where $1\leq i\leq R$, by computing the median value of the factors in each cluster~(Step \textcircled{4}) and form the consensus matrix:
\begin{equation}
  \overline{A^{(k)}}_{C}=[ f^{(k)}_{C_1} f^{(k)}_{C_2} \ldots f^{(k)}_{C_{R}} ] \in \mathbb{R}^{I_k \times R}, 1\leq k \leq N.
\end{equation}
We perform the decomposition using $\overline{A^{(k)}}_{C}$ as the fixed initial guess for the $k$-th modality to obtain the final factor matrices (Step \textcircled{5}). 

Notice that if ZIPTF is employed as the factorization method in Step \textcircled{1} described above, we refer to the resulting factorization as C-ZIPTF. 
\section{Experiments}\label{section:experiments}
Here we present results showing the superior performance of C-ZIPTF across multiple evaluation metrics. We implemented C-ZIPTF in Python, using the probabilistic programming language Pyro \cite{Pyro}. Our presentation in Section \ref{BPF} focused on Bayesian tensor factorization frameworks that utilize Poisson and Zero-Inflated Poisson based models. However, our implementation is designed to be more versatile and can accommodate different types of noise models. 
We conduct three different evaluations to assess the performance of our proposed method. First, we compare ZIPTF with alternative factorization methods on simulated tensors with known factors and ZIP noise and evaluate the benefits of using a ZIP model and the inclusion of the consensus approach (Section \ref{section:synthetic_tensor}). Second, we evaluate the performance of our method on simulated single-cell RNA sequencing data and compare it with other matrix and tensor factorization methods at the task of recovering identity and activity gene expression programs (GEPs), Section~\ref{section:simulated}. Finally, we demonstrate the ability of our method to capture biologically meaningful gene expression programs by applying it to a real single-cell RNA sequencing dataset of immune cells stimulated with interferon beta (Section \ref{section:single-cell}).
\begin{centering}
\begin{figure*}[t]\label{fig:synthetic}
    \includegraphics[width=\linewidth]{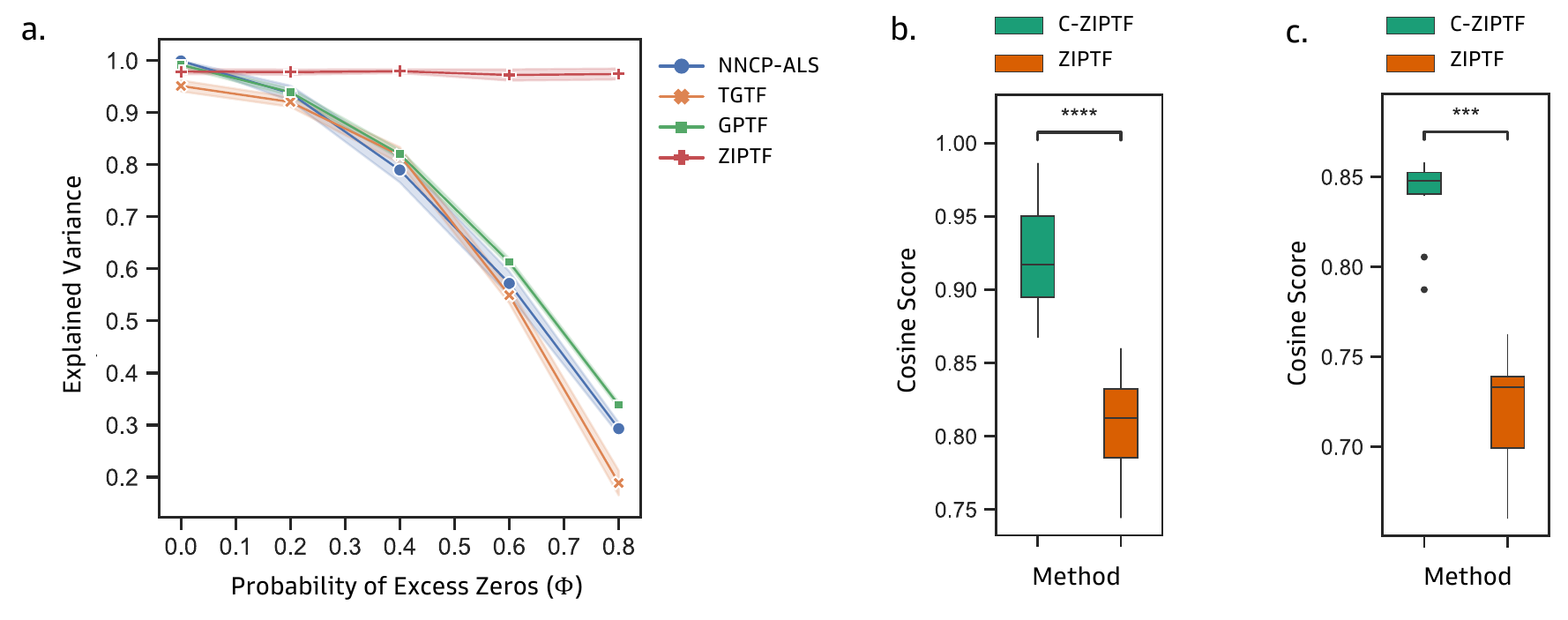}
  \caption{ZIPTF compared to alternative factorization methods on a synthetic tensor with known factors and ZIP noise, and stability comparison between ZIPTF and C-ZIPTF. (a) We calculated the explained variance of ZIPTF and alternative methods for different levels of extra zeros.
   (b) Cosine similarity between factors obtained on repeat runs for ZIPTF and C-ZIPTF. (c) Cosine similarity between inferred factors and original factors for ZIPT and C-ZIPT.}
   \label{fig:synthetic}
\end{figure*}
\end{centering}
\subsection{Synthetic tensor experiment}\label{section:synthetic_tensor}
To evaluate the performance of C-ZIPTF on synthetic data, we generate tensors using known factors and Poisson noise with varying degrees of zero inflation and measure the accuracy of different methods at recovering the original factors. 
To generate a tensor $\mathcal{T'}\in \mathbb{R}^{I\times J \times K}$, we first create three factor matrices $A\in \mathbb{R}_{+}^{I\times R}$, $B\in \mathbb{R}_{+}^{J\times R}$, and $C\in \mathbb{R}_{+}^{K\times R}$, with elements drawn from a Gamma distribution with shape $\alpha=3$ and rate $\beta=0.3$, where $R$ is the desired true rank. We then construct a tensor $\mathcal{T}$ by taking the sum of the outer product of the corresponding columns of the matrices, i.e., $\mathcal{T} = [[A,B,C]]$. Finally, we generate the tensors $\mathcal{T}'$ by sampling from a ZIP distribution with mean $\mathcal{T}$ and a given probability of extra zeros, denoted by $\Phi$ in Section \ref{section:ZIPTF}. 
\vspace*{1\baselineskip} 

\noindent\textbf{Zero-inflated Poisson model results in higher explained variance}
For the first experiment, we ran ZIPTF without consensus aggregation to evaluate the advantages of using the ZIP model alone. For comparison, we considered three alternative tensor factorization methods: Non-Negative CP decomposition via Alternating-Least Squares~(NNCP-ALS) \cite{Kolda}, Bayesian Tensor Factorization with Truncated Gaussian model~(TGTF)\cite{probabilistic}, and Bayesian Tensor Factorization with Gamma Poisson model~(GPTF)\cite{bayesian}. We conducted 20 trials, generating a new simulated tensor $\mathcal{T}'$ of shape $10 \times 20 \times 300$ and rank 9 each time and running each factorization method on the tensor for a fixed maximum number of iterations ($max\_iter = 1000$). We evaluated the performance of the methods using the explained variance  \eqref{explained-variance} of the approximation generated by each factorization.
ZIPTF consistently outperformed both NNCP-ALS and the Bayesian Tensor Factorization methods without the ZIP model, as shown in Figure \ref{fig:synthetic}(a). At a zero probability of excess zeros, all four methods showed similar and nearly perfect explained variance. However, as the excess zero level increased, the performance of the other methods deteriorated rapidly.
At the highest probability of excess zeros simulated, $\Phi = 0.8$, the average explained variance of the ZIPTF approximation was 0.974, with a 95\% confidence interval (CI) [0.962, 0.987], about $2.4\times$ better than the second highest explained variance of 0.338, 95\% CI [0.334, 0.342] achieved by the Gamma Poisson model. We also note that the difference in explained variance between NNCP-ALS and the Bayesian methods other than ZIPTF is minimal compared to the difference to ZIPTF. This indicates that the superiority of ZIPTF arises from using the appropriate noise model. 
\vspace*{1\baselineskip} 

\noindent \textbf{Consensus aggregation leads to more consistent factorization}
After demonstrating ZIPTF's superior performance in modeling zero-inflated count data, we examine the benefits of consensus aggregation. We generate tensors of shape $40 \times 20 \times 2000$ and rank 9 with known factors and Zero-Inflated Poisson noise as described above and evaluate the recovered factors by running ZIPTF with and without consensus aggregation. For this experiment, we fix the probability of excess zeros $\Phi = 0.6$.
We compare the internal consistency of factors obtained from multiple runs of the decompositions. For our simulated tensor $\mathcal{T}',$ assume that we have two rank R approximations $[[A,B,C]]$ and $[[D,E,F]]$ corresponding to different randomly initialized runs. To measure the similarity between factorizations, we calculate:
\begin{equation}\label{cosine_similarity}
\textit{cosine score}([[A,B,C]], [[D,E,F]])= \frac{1}{R}\sum_{i=1}^{R} \max_{1\leq j \leq R}~ \cos(a_i, d_j)\cos(b_i,e_j) \cos(c_i, f_j). \
\end{equation}
We evaluate the similarity of factors recovered from 20 randomly initialized runs of both ZIPTF and C-ZIPTF using the \textit{cosine score} given in Eqn. \ref{cosine_similarity}. We observe that the factors recovered from C-ZIPTF are more consistent with one another compared to those recovered from ZIPTF, as shown in Figure \ref{fig:synthetic}(b). The consensus approach makes C-ZIPTF more robust, reducing the impact of the inherent stochasticity of the factorization process and resulting in a more stable set of factors.
\vspace*{1\baselineskip} 

\noindent \textbf{Consensus aggregation leads to more accurate recovery of original factors}
We assess the accuracy of both ZIPTF and C-ZIPTF in recovering the original factors used to create the tensor with $\Phi=0.6$. We perform 20 randomly initialized runs of each method and compare the recovered factors to the original factors using the \textit{cosine score}. Figure 2(c) demonstrates that C-ZIPTF outperforms ZIPTF in recovering the original factors.
\subsection{Synthetic single-cell RNA-Seq data analysis}\label{section:simulated}

We test the performance of C-ZIPTF on single-cell RNA sequencing (scRNA-seq) data, which is prone to zero inflation due to technical limitations that result in dropout events \cite{RNA-seq-challanges}.
To evaluate the effectiveness of C-ZIPTF, we compared its performance with other matrix and tensor factorization methods using a synthetic scRNA-seq dataset. We used the Splatter simulation framework \cite{Splatter} which was adapted to Python in a previous study \cite{consensus-nmf} to generate the synthetic scRNA-seq dataset.

\begin{centering}
\begin{figure*}[h]
    \includegraphics[width=\linewidth]{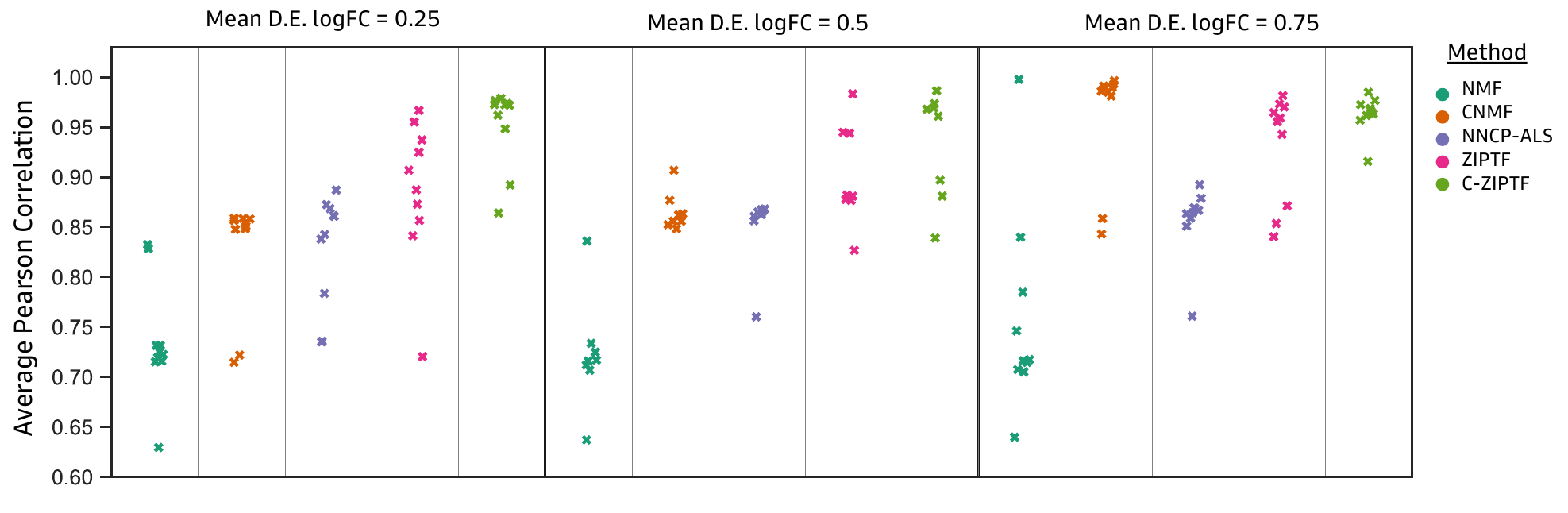}
  \caption{The average Pearson correlation between the true gene expression programs (GEPs) used in the simulation and the inferred GEPs obtained from various factorization methods. The results are presented for three different signal intensity levels (0.25, 0.5, 0.75), which are indicated by the mean log2 fold change (log2FC) of simulated differentially expressed genes.}
  \label{fig:simulated_single_cell}
\end{figure*}
\end{centering}
The simulation framework utilizes a Gamma-Poisson hierarchical model with hyper-parameters estimated from real data. Technical dropouts are simulated by randomly replacing some of the simulated counts with zeros using a Bernoulli distribution. The complete details of the simulation framework and parameters used are provided in Section~\ref{section:supplementary}. The synthetic dataset consists of 15,000 cells and 5,000 genes from six donors, with five gene expression programs defining cell type identities and three gene expression programs defining donor-specific activity. We evaluated the performance of various factorization methods, including ZIPTF, C-ZIPTF, Non-negative Matrix Factorization (NMF) \cite{NMF-Lee}, Consensus NMF (CNMF) \cite{consensus-nmf}, and NNCP-ALS, with the goal of recovering the eight gene expression programs embedded in the synthetic scRNA-seq dataset.
For NMF and CNMF, the rank 8 decomposition is performed with a maximum of 1000 iterations for convergence after normalizing the cell-by-gene count matrix to counts per million (CPM).
For the tensor-based approaches, we construct the observed tensor by pseudobulking the cell-by-gene counts matrix. We cluster the cells to obtain tentative \emph{cell type} groupings and generate pseudobulk counts by summing all the counts for each donor, cell type, and gene. This creates a tensor of shape $D \times C \times G$, where D, C, and G represent the number of donors, cell types, and genes, respectively. We then normalize the pseudobulk tensor to CPM and apply tensor factorization methods with rank 8 and perform 1000 iterations for each method.

To evaluate the performance of the factorization methods in recovering the eight true gene expression programs (GEPs), we computed the Pearson correlation \cite{Pearson} between each of the eight latent factors in the gene mode obtained via factorization and the original GEPs. This correlation was used to establish a one-to-one alignment between the factors and the GEPs. We calculated the average Pearson Correlation between each factor and its corresponding GEP as the accuracy score of the method. The results of our analysis are presented in Figure \ref{fig:simulated_single_cell} for three different levels of simulated intensity for the activity GEPs~(mean log2 fold change of differentially expressed genes (log2FC) $ \in \{0.25, 0.5, 0.75 \}$). We observed that when the signal is strong enough (log2FC $ = 0.75 $), CNMF and C-ZIPTF perform comparably. However, when the signal intensities are lower (log2FC $\in \{0.25, 0.5\}$), C-ZIPTF clearly outperforms all other methods. Additionally, we found that ZIPTF without consensus aggregation also performs better than the other factorization methods, indicating that both the use of the ZIP model and the consensus aggregation improve the accuracy of the method in recovering GEPs. 
  \label{fig:real_single_cell}
\subsection{Real single-cell RNA-Seq data analysis}\label{section:single-cell} 
\begin{figure*}[t!]
\includegraphics[width=\linewidth]{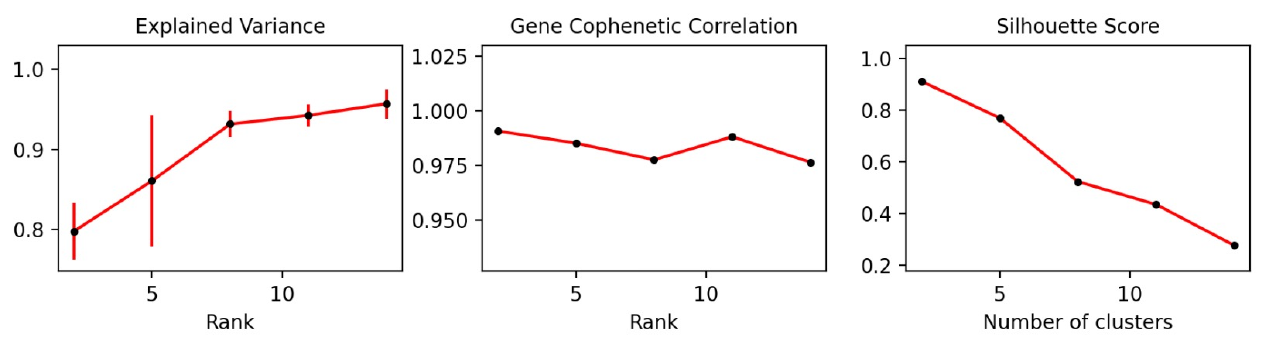}
  \caption{Metrics used in selecting the optimal rank for running C-ZIPTF on real single-cell RNA sequencing
dataset \cite{PBMCS} of immune cells stimulated with interferon beta (IFN-$\beta$).}
   \label{fig:rank_metrics}
\end{figure*}

We applied C-ZIPTF to a real-world single-cell RNA sequencing dataset of peripheral blood mononuclear cells (PBMCs) from patients with Lupus, reported in \cite{PBMCS}.
We obtained the single-cell RNA-Seq data from GEO using accession number GSE96583 \url{https://www.ncbi.nlm.nih.gov/geo/query/acc.cgi?acc=GSE96583}. As described in \cite{PBMCS}, the dataset contains 29,065 cells from eight patients, which are divided into stimulated and control groups, with the former being treated with interferon beta (IFN-$\beta$), a cytokine that modulates the transcriptional profile of immune cells.  As part of the preprocessing step, we filter out multiplets and cells without a cell type assignment. Additionally, we remove samples and cell types that constitute less than 2 percent of cells. After these filtering steps, the dataset contained 14 samples, 7 control and 7 stimulated, and 6 cell types: CD4 T-cells, CD14+ Monocytes, B-cells, CD8 T-cells, NK- cells, and FCGR3A+ Monocytes. In order to facilitate biological interpretability of factors and reduce noise in the tensor formed we removed genes that are either not provided with HGNC symbols \cite{HGNC}, or had a total count of less than 50 across all cells. Finally, we create a pseudobulk tensor by summing up the raw counts for each cell type, sample, and gene. The resulting pseudobulk data tensor has dimensions S $\times$ C $\times$ G  (14 $\times$ 6 $\times$ 9,276), where S, C and G denote the number of samples, cell types and genes respectively. We normalize the tensor such that each sample-cell type pair has a total of $10^6$ counts. We first determined the optimal rank for the data by running C-ZIPTF with a range of ranks from 2 to 14 and just 5 restarts. Some of the metrics we considered in deciding the optimal rank including explained variance and silhouette score are shown in Figure \ref{fig:rank_metrics}. We select rank 8, which exhibited a high explained variance of 0.969.

\begin{figure*}[t!]
\includegraphics[width=\linewidth]{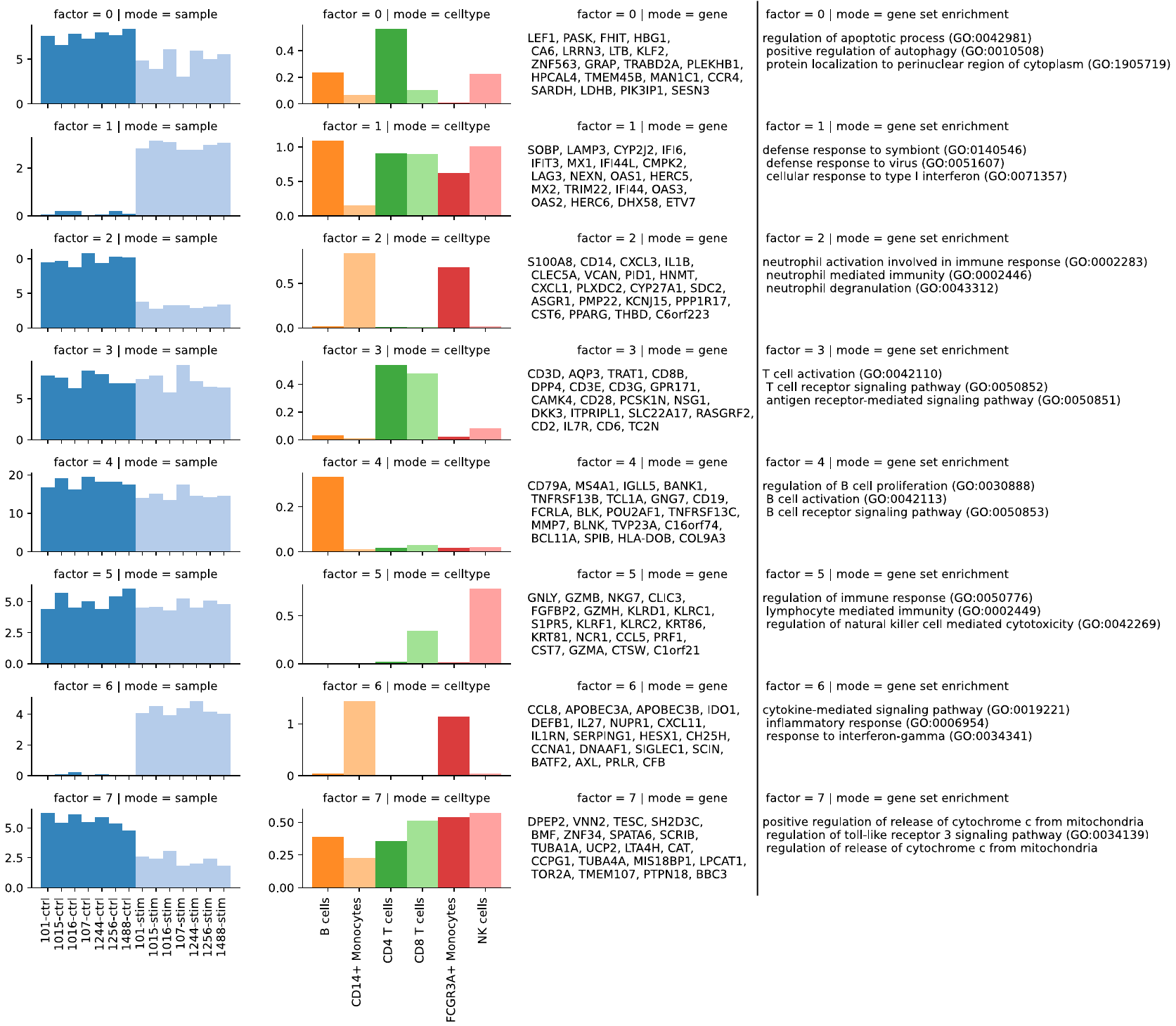}
  \caption{Full set of factors recovered by running C-ZIPTF on real single-cell RNA sequencing
dataset \cite{PBMCS} of immune cells stimulated with interferon beta (IFN-$\beta$). Each row represents a factor,
and the first three columns display the three modes: sample, cell type, and gene. The $y-$axis in the
sample and cell type modes represent the loading of the sample or cell type on that factor. The
gene mode exhibits the top 20 genes associated with the factor. The last column provides the top 3
enriched terms obtained from a gene set enrichment analysis.}
   \label{fig:full_fact}
\end{figure*}

As depicted in Figure \ref{fig:full_fact}, C-ZIPTF successfully identifies both cell type identity and condition-specific gene expression programs (GEPs). Notably, factor 4 represents an identity GEP that remains active in all B-cells, irrespective of the condition. The genes exhibiting the highest loadings for this factor are well-established B-cell markers, such as \textit{MS4A1}, \textit{CD79A}, and \textit{BANK1} \cite{B-cells}. Furthermore, we performed gene set enrichment analysis~\cite{gene-set-enrichment} of these factors using GSEApy~\cite{GSEApy} in Python. This analysis revealed enrichment pathways consistent with B-cell characteristics, including B-cell activation and the B-cell receptor signaling pathway.

Conversely, factor 1 and factor 6 capture distinct gene expression programs that are specifically activated in IFN-$\beta$ stimulated samples. Factor 1 captures a cross-cell-type response to IFN-$\beta$ stimulation, whereas factor 6 represents a Monocyte-specific response. These findings align with previous studies that have reported a Monocyte-specific response to IFN-$\beta$ stimulation \cite{Lotfollahi}. Furthermore, gene set enrichment analysis revealed enrichment in pathways such as the cellular response to type I interferon and inflammatory response, among others. 
For a comprehensive list of factors identified by C-ZIPTF and their associated gene expression programs, please refer to Figure \ref{fig:full_fact}.


\section{Conclusion}
Zero-inflated count data is a common phenomenon in a wide range of fields, including genomics, finance, risk management, healthcare, and social sciences. However, traditional tensor factorization methods have limited effectiveness when dealing with zero-inflated data, often yielding inaccurate and unstable results across runs with different initializations. To overcome these challenges, we propose ZIPTF, a Bayesian tensor factorization model that is specifically tailored to zero-inflated count data. Additionally, we introduce a generic meta-analysis framework for consensus-driven tensor factorization. By combining these two approaches, we develop a novel method called C-ZIPTF that achieves both high accuracy and stability, and outperforms state-of-the-art baselines on synthetic and real data. Our proposed method provides a useful tool for researchers in various fields to gain deeper insights into their data.

The Bayesian approach for tensor factorizations offers several other advantages over maximum likelihood estimation-based methods for tensor factorization. These include the ability to incorporate prior knowledge, perform model selection, and quantify uncertainty in the parameter estimates. However, it is important to note that Bayesian methods can be computationally expensive and require careful specification of prior distributions, which may require expert knowledge. Moreover, the tensor methods discussed in this paper rely on a multilinear factorization form and may be inadequate for capturing more complex, nonlinear relations in the data. To overcome this limitation, one possible solution is to integrate a kernelized approach into the factorization. In future work, we plan to focus on the careful design of kernel functions that would enable us to effectively capture nonlinear patterns in the data. 

\section{Supplementary Material}\label{section:supplementary}

\subsection{Implementation of ZIPTF and C-ZIPTF}\label{subsec:impl_cziptf}\label{imp1}

We present a Python implementation of a versatile Bayesian Tensor Factorization method using Variational Inference. Our implementation leverages Pyro \cite{Pyro}, a probabilistic programming framework built on PyTorch. The BayesianCP class inherits from \texttt{torch.nn.Module} and offers functionalities for model fitting and summarizing the posterior distribution of factor matrices. During model fitting, Stochastic Variational Inference (SVI) is employed with an Adam optimizer \cite{Adam,SVI}. The current implementation supports three models: Zero Inflated Poisson model (ZIPTF), a Gamma Poisson model (GPTF) \cite{bayesian}, and a Truncated Gaussian model (TGTF) \cite{probabilistic}.

\subsection{Implementation of baseline methods}\label{subsec:base_impl}
As mentioned in Section \ref{imp1}, we utilize the same implementation for the other Bayesian tensor factorization approaches (Gamma Poisson Bayesian Tensor Factorization and Truncated Gaussian Bayesian Tensor Factorization) as the ZIPTF method and the code is provided at  \url{https://github.com/klarman-cell-observatory/scBTF} and \url{https://github.com/klarman-cell-observatory/scbtf_experiments}. For the remaining baselines used in our comparisons we use the following implementations:

\begin{itemize}
  \item \textbf{Non-negative Matrix Factorization (NMF)}: We use the Python implementation provided in the scikit-learn package. \url{https://scikit-learn.org/stable/modules/generated/sklearn.decomposition.non_negative_factorization.html}.
  \item \textbf{Consensus Non-negative Matrix Factorization (cNMF)}: We use the Python implementation described in \cite{consensus-nmf}, and provided on GitHub \url{https://github.com/dylkot/cNMF/tree/master}
  \item \textbf{Non-negative CP via Alternating-Least Squares (NNCP-ALS)}: We use the Python implementation provided in the Tensorly package. \url{http://tensorly.org/stable/modules/generated/tensorly.decomposition.non_negative_parafac_hals.html}
\end{itemize}

\subsection{Simulation details}{}

We use a Python adaptation of the Splatter \cite{Splatter} statistical framework given in \cite{consensus-nmf} to simulate single-cell RNA-Seq data. The core of the simulation is a Gamma-Poisson distribution used to generate a cell-by-gene count matrix. While the original Splatter framework supports the simulation of both expression outlier genes and technical dropout (random knockout of counts), the Python adaptation in \cite{consensus-nmf} only keeps outlier expression simulation. Since our method is specifically adapted to handle dropout noise in single-cell data, we add back the modeling of dropout to the Python adaptation. Specifically, after sampling counts from a Poisson distribution, we simulate dropout noise by calculating the probability of a zero for each gene from its mean expression and using that to randomly replace some of the simulated counts with zeros employing a Bernoulli distribution as described in \cite{Splatter}.

The distribution of expression values prior to incorporating differential expression was determined based on parameters estimated from a random sample of 8000 cells from an organoid dataset as described in \cite{consensus-nmf}. Specifically, the library size of a cell is sampled from a Lognormal distribution derived from a Normal distribution with a mean of 7.64 and a standard deviation of 0.78. The mean expression of a gene is sampled from a Gamma distribution with a mean of 7.68 and a shape of 0.34. With the probability of 0.00286, a gene will be an outlier from this Gamma distribution and will instead be sampled from a Lognormal distribution derived from a Normal distribution with a mean of 6.15 and standard deviation of 0.49. Additionally, we set a 5\% doublet rate. Doublets are formed by randomly sampling a pair of cells, combining their gene counts, and downsampling such that the total count equals the larger of the two.

\newpage


\begin{thebibliography}{1}
\bibitem{Acar} E. Acar, T. Kolda, D. M. Dunlavy, {\it All-at-once optimization for coupled matrix and tensor factorizations}, arXiv:1105.3422 (2011).







\bibitem{Pearson} J. Benesty, J. Chen, Y. Huang, and I. Cohen, {\it Pearson correlation coefficient}, Noise reduction in speech processing (2009), pp. 37–40, Springer.


\bibitem{Pyro} E. Bingham, J.P. Chen, M. Jankowiak, F. Obermeyer, N. Pradhan, T. Karaletsos, R. Singh, P. Szerlip, P. Horsfall, and N.D. Goodman, N.D, {\it Pyro: Deep Universal Probabilistic Programming}, Journal of Machine Learning Research (2018), 19(108) , 1-6.




\bibitem{Bishop} C. Bishop, {\it Pattern Recognition and Machine Learning}, New York:Springer (2006).


\bibitem{David}D. M. Blei, A. Kucukelbir, and J. D. McAuliffe, {\it Variational Inference: A Review for Statisticians}, Journal of the American Statistical Association (2017), 112:518, 859-877, DOI: 10.1080/01621459.2017.1285773

\bibitem{LOF} M.M. Breunig, H.-P.Kriegel, R.T. Ng, and J. Sander, {\it LOF: Identifying density-based local outliers,} Proceedings of the 2000 ACM SIGMOD international conference on Management of data—SIGMOD ‘00, Dallas, TX, USA.

\bibitem{Brunet}J.P. Brunet, P. Tamayo, T.R. Golub, J.P. Mesirov, and E. S. Lander,~{\it Metagenes and Molecular Pattern Discovery Using Matrix Factorization}, Proceedings of the National Academy of Sciences of the United States of America (2004), 101(12), 4164–4169, \url{http://www.jstor.org/stable/3371580}.



\bibitem{Cemgil} A. Cemgil, {\it Bayesian inference for nonnegative matrix factorisation models}, Computational Intelligence and Neuroscience (2009).

\bibitem{nmf2} J.C. Chen, {\it The nonnegative rank factorizations of nonnegative matrices},~Linear Algebra Appl.(1984), vol. 62, pp. 207–217.

\bibitem{tensor-sparse} E.C. Chi and T.G. Kolda, {\it On tensors, sparsity, and nonnegative factorizations}, SIAM Journal on Matrix Analysis and Applications~(2012), 33 1272–1299.

\bibitem{Chigona}M. Chigona and C. Gaetan, {\it Semiparametric zero-inflated Poisson models with applications to animal abundance studies},~Environmetrics (2007), 18(3), 303-314.





\bibitem{B-cells} C. Domínguez Conde, C. Xu, L.B. Jarvis, and others, {\it Cross-tissue immune cell analysis reveals tissue-specific features in humans}, Science(2022), 376(6594), eabl5197.



\bibitem{GSEApy} Z. Fang, X. Liu, and G. Peltz, {\it GSEApy: a comprehensive package for performing gene set enrichment analysis in Python}, Bioinformatics, Volume 39, Issue 1, January (2023), btac757, \url{https://doi.org/10.1093/bioinformatics/btac757}





\bibitem{recommender1} E. Frolovand I.  Oseledets, {\it Tensor methods and recommender systems}, WIREs Data Mining Knowl Discov (2017), 7: e1201, \url{https://doi.org/10.1002/widm.1201}.

\bibitem{Ghosh} S.K. Ghosh, P Mukhopadhyay, and J.C. Lu, {\it  Bayesian analysis of zero-inflated regression models}, Journal of Statistical Planning and Inference (2006), 136(4), 1360-1375.



\bibitem{poisson-nmf-recommender} P. Gopalan, J. Hofman, and D. Blei, {\it Scalable recommendation with Poisson factorization}, Proceedings of the Thirty-First Conference on Uncertainty in Artificial Intelligence (2015).



\bibitem{probabilistic} J. L. Hinrich, K. H. Madsen, and M. Mørup, {\it The probabilistic tensor decomposition toolbox},~Machine Learning: Science and Technology (2020), available online at \url{https://github.com/JesperLH/prob-tensor-toolbox}.

\bibitem{SVI} M.D. Hoffman, D.M. Blei, C. Wang, and J. Paisley,{\it Stochastic variational inference}(2013), Journal of Machine Learning Research.

\bibitem{BNTF-3} C. Hu, P. Rai, C. Chen, M. Harding, and L Carin, {\it  Scalable Bayesian Non-negative Tensor Factorization for Massive Count Data}, Machine Learning and Knowledge Discovery in Databases (2015), ECML PKDD.





\bibitem{tensor-survey} Y. Ji, Q. Wang, X. Li, and J. Liu, {\it A Survey on Tensor Techniques and Applications in Machine Learning},~IEEE Access(2019), vol. 7, pp. 162950-162990, doi: 10.1109/ACCESS.2019.2949814.

\bibitem{Monti} I. Jung, M. Kim, S. Rhee, S. Lim, and S. Kim, {\it MONTI: A Multi-Omics Non-negative Tensor Decomposition Framework for Gene-Level Integrative Analysis}, Frontiers in genetics (2021), 12, 682841. \url{https://doi.org/10.3389/fgene.2021.682841}.

\bibitem{PBMCS} H.M Kang, M. Subramaniam,~and~others, {\it Multiplexed droplet single-cell RNA-sequencing using natural genetic variation}, Nature biotechnology (2018), 36(1), 89–94, \url{https://doi.org/10.1038/nbt.4042}.


\bibitem{Adam} D.P. Kingma and J. Ba, {\it Adam: A method for stochastic optimization} (2014), arXiv preprint arXiv:1412.6980.


\bibitem{Kolda}T.~G.~Kolda and B.~W.~Bader, {\it Tensor decompositions and applications}, SIAM review~{\bf 51}~(2009), no.~3, 455--500.



\bibitem{consensus-nmf}D. Kotliar, A. Veres, M. A. Nagy, S. Tabrizi, E. Hodis, D.A. Melton, and P.C. Sabeti, {\it Identifying gene expression programs of cell-type identity and cellular activity with single-cell RNA-Seq}, eLife (2019), 8, e43803, \url{https://doi.org/10.7554/eLife.43803}



\bibitem{Kurt}B. Kurt, {\it Bayesian poison tensor factorization for dummies} (2018), \url{https://github.com/bariskurt/bptf}.

\bibitem{Lambert} D. Lambert, {\it Zero-Inflated Poisson Regression, with an Application to Defects in Manufacturing}. Technometrics (1992), 34(1):1-14.

\bibitem{RNA-seq-challanges}D. Lähnemann, J. Köster, E. Szczurek, D. J. McCarthy, S. C. Hicks, M. D. Robinson, C. A. Vallejos, K. R. Campbell, and  others, {\it Eleven grand challenges in single-cell data science}, Genome biology (2020), 21, 1--35.

\bibitem{NMF-Lee} D. D. Lee and H. S. Seung, {\it Learning the parts of objects by non-negative matrix factorization}, Nature (1999), vol. 401, no. 6755, pp. 788–791.


\bibitem{Lotfollahi} M. Lotfollahi, S. Rybakov, and others, {\it Biologically informed deep learning to query gene programs in single-cell atlases}, Nat Cell Biol (2023), 25, 337–350,\url{https://doi.org/10.1038/s41556-022-01072-x}


\bibitem{K-means}J. MacQueen, {\it Some methods for classification and analysis of multivariate observations}~(1967).



\bibitem{denoising} A. Özdemir, M. A. Iwen, and S. Aviyente, {\it A multiscale approach for tensor denoising}, IEEE Statistical Signal Processing Workshop (2016), pages 1–5. 





\bibitem{Bayesian-inference-matrix} J. Paisley, D. Blei, and M. Jordan, {\it Bayesian nonnegative matrix factorization with stochastic variational inference}, Handbook of Mixed Membership Models and Their Applications (2014). Chapman and Hall/CRC.






\bibitem{black-box}R. Ranganath, S. Gerrish, and D. Blei,  {\it Black box variational inference} (2014),  Artificial intelligence and statistics, PMLR (pp. 814-822)

\bibitem{Robbins} H. Robbins and S. Monro,{\it A stochastic approximation method}(1951), The annals of mathematical statistics, 400-407.



\bibitem{silhouette} P.J. Rousseeuw,~{\it Silhouettes: a graphical aid to the interpretation and validation of cluster analysis}, Journal of computational and applied mathematics (1987), 20, 53-65.



\bibitem{bayesian} A. Schein, J.  Paisley, D. Blei, David, and H. Wallach, {\it Bayesian Poisson Tensor Factorization for Inferring Multilateral Relations from Sparse Dyadic Event Counts},~Proceedings of the 21st ACM SIGKDD International Conference on Knowledge Discovery and Data Mining (2015), \url{https://doi.org/10.1145/2783258.278341410.1145/2783258.2783414}. 


\bibitem{HGNC} R.L. Seal, B. Braschi, and others, {\it Genenames.org: the HGNC resources in 2023}, Nucleic Acids Res. PMID: 36243972, \url{DOI: 10.1093/nar/gkac888}.









\bibitem{Simchowitz} M. Simchowitz, {\it Zero-inflated Poisson factorization for recommendation systems}, Junior Independent Work (advised by D. Blei), Princeton University, Department of Mathematics (2013).


\bibitem{gene-set-enrichment} A. Subramanian, P. Mootha, and others,{ \it Gene set enrichment analysis: a knowledge-based approach for interpreting genome-wide expression profiles}, Proceedings of the National Academy of Sciences~(2005), 102(43), 15545-15550.



\bibitem{Bro} G. Tomasi and R. Bro, {\it A comparison of algorithms for fitting the parafac model}, Comput. Stat. Data Anal.~(2006),50 1700–34.




\bibitem{semi-nonnegative} M. Wang, J. Fischer, and  Y. S. Song, {\it Three-way clustering of multi-tissue multi-individual gene expression data using semi non-negative tensor decomposition}, The annals of applied statistics (2019), 13(2), 1103–1127, \url{https://doi.org/10.1214/18-aoas1228}


\bibitem{NTF-image1} H. Yang, G. He, and Y. Dong, {\it Nonnegative Tensor Decomposition and It's Applications in Image Processing}, Proceedings of the International Conference on Computer Science and Information Technology, ICCSIT (2008), 212 - 217. 10.1109/ICCSIT.2008.99.

\bibitem{BTF-cemgil2} Y.K. Yılmaz and A.T. Cemgil, {\it Probabilistic Latent Tensor Factorization} in {\it Latent Variable Analysis and Signal Separation}, Lecture Notes in Computer Science (2010), vol 6365. Springer, Berlin, Heidelberg, \url{https://doi.org/10.1007/978-3-642-15995-4\_43}.


\bibitem{Splatter} L. Zappia, B. Phipson, and A. Oshlack, {\it Splatter: simulation of single-cell RNA sequencing data}, Genome Biol (2017),18 174, \url{https://doi.org/10.1186/s13059-017-1305-0}.

\bibitem{beta} M. Zhou, L. Hannah, D. Dunson, and L. Carin,~{\it Beta-negative binomial process and Poisson factor analysis}. arXiv:1112.3605 (2011).









\end{thebibliography}
\end{document}